\documentclass[journal,10pt,twocolumn]{IEEEtran}

\usepackage{cite}
\usepackage{amsmath,amssymb,amsfonts}
\usepackage{algorithmic}
\usepackage{graphicx}
\usepackage{textcomp}
\usepackage{xcolor}
\usepackage{booktabs}

\begin{document}

\title{Error-Aware TF-IDF Retrieval-Augmented Generation for ASR Error Correction}

\author{Mohammad Aref Jafari-Raddani%
\thanks{Mohammad Aref Jafari-Raddani is with the Department of Computer Engineering, Qom University of Technology, Qom, Iran, and also with Asa Electronic Akhtaran, Isfahan, Iran (e-mail: raddaniaref@gmail.com).}}

\maketitle

\begin{abstract}
End-to-end automatic speech recognition systems frequently hallucinate rare entities and domain-specific terms, especially in low-resource languages. While retrieval-augmented generation frameworks can mitigate these errors using large language models, current architectures face significant challenges. They either rely on standard sparse retrieval that ignores phonetic misrecognitions or utilize heavyweight cross-modal embeddings that introduce high latency. This letter proposes a highly efficient, purely lexical error-aware framework designed to explicitly resolve phonetic and loop hallucinations. Our approach integrates a symmetric text normalization module with a novel error-aware term frequency-inverse document frequency algorithm. By constructing a sparse diagonal penalty matrix based on historical errors, the retriever mathematically prioritizes corrective documents containing specific high-risk misrecognitions. Evaluated on the Persian subset of the FLEURS dataset, our method increased the error-aware hit rate from 53.7\% to 90.9\%. In end-to-end evaluations, the integrated framework reduced the final word error rate from 23.06\% to 18.83\%, achieving significant accuracy gains with near-zero inference latency.
\end{abstract}


\begin{figure*}[htbp]
    \centering
    \includegraphics[width=\linewidth]{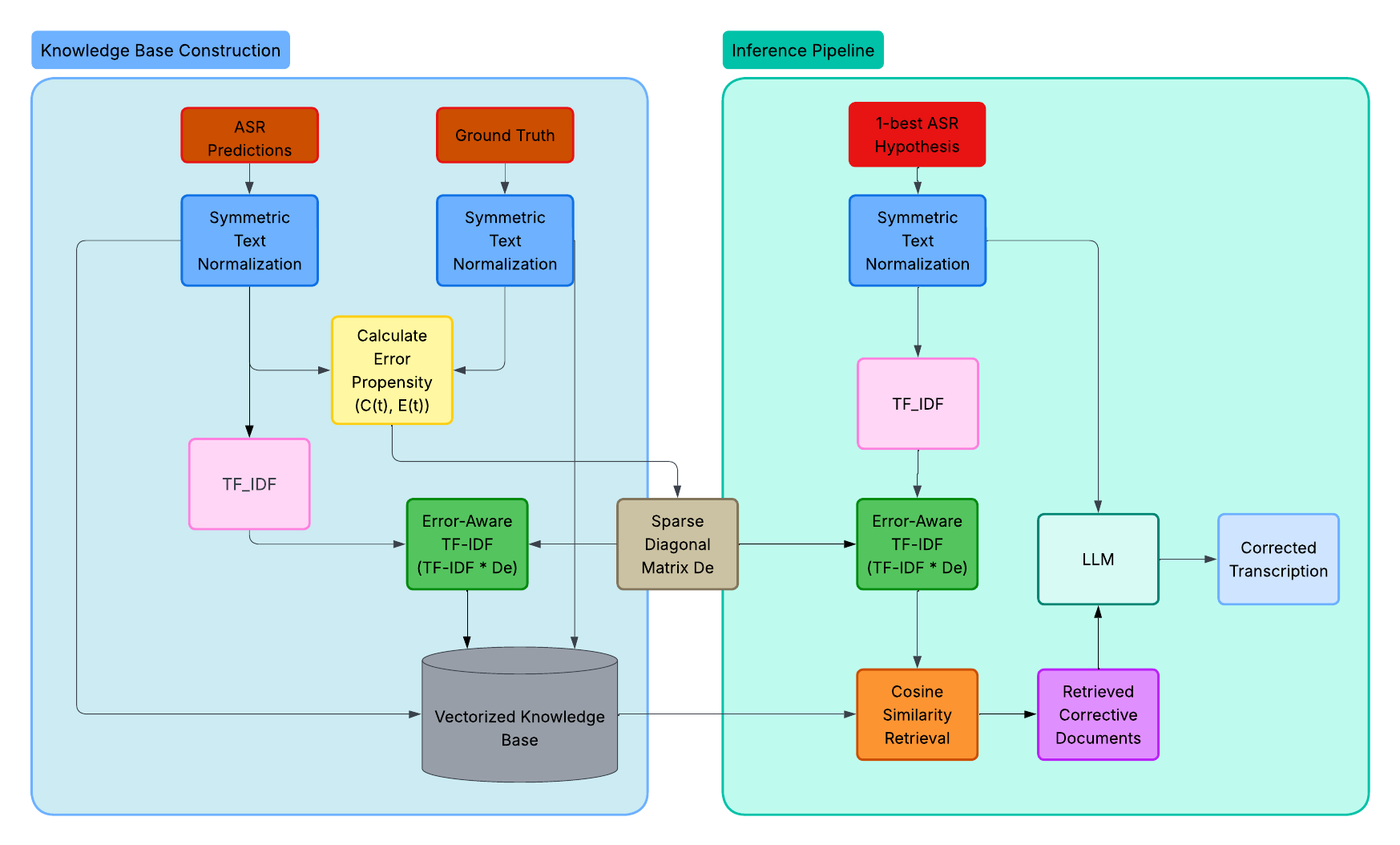}
    \caption{Overall architecture of our proposed framework for Knowledge Base (KB) construction and inference. While sharing structural similarities with standard ASR-RAG paradigms \cite{robatian2025gec}, we introduce two computationally free modifications: parallel preprocessing via Symmetric Text Normalization and a novel Error-Aware TF-IDF retrieval algorithm. These enhance retrieval performance without adding inference latency.}
    \label{fig:architecture}
\end{figure*}

\section{Introduction}

\IEEEPARstart{A}{utomatic} 
Speech Recognition (ASR) plays a vital role in human-computer interaction by providing foundational speech-to-text transcription \cite{rebman2003speech}. In recent years, ASR has been revolutionized by the advent of end-to-end deep learning models \cite{prabhavalkar2023end}. Nevertheless, in low-resource languages, the performance drops significantly due to environmental noise \cite{chen2022noise} and a wide variety of accents \cite{turan2022adapting}. In addition, during real-world usage in specific domains, these models consistently struggle to recognize rare named entities \cite{wang2024dancer}.

To mitigate these limitations, a wide range of post-processing methods have been proposed. With the advancement of Large Language Models (LLMs), utilizing LLMs for post-ASR error correction has become a major area of study, commonly known as Generative Error Correction \cite{ma2023can, chen2023hyporadise}. For instance, Ma et al. \cite{ma2023can} demonstrated how the error rate of ASR can be reduced by leveraging LLMs for error correction, while Gu et al. \cite{gu2024denoising} improved ASR accuracy through LLM-based contextual spelling correction. Despite their deep linguistic comprehension, LLMs are susceptible to hallucination, especially when resolving unknown or rare named entities that fall outside their pre-trained parameters. To address this, recent studies have explored providing LLMs with augmented data to enable more focused generative correction. Pusateri et al. \cite{pusateri2025retrieval} used a vector database of entities and an LLM for correcting named entity errors. Furthermore, Robatian et al. \cite{robatian2025gec} introduced GEC-RAG, which uses a TF-IDF vector database of ASR predictions and ground truths to provide the LLM with similar mistakes and their corrected versions.

Recognizing the limitations of standard lexical retrieval, recent state-of-the-art methodologies have pivoted towards highly complex, cross-modal retrieval architectures. For example, recent advancements employ Finite Scalar Quantization (FSQ) to compress massive contextual cross-attention embeddings \cite{flemotomos2025optimizing}, or utilize speech-and-bias contrastive learning to directly align acoustic and textual modalities at scale \cite{gong2025br}. Moreover, other approaches attempt to explicitly map phonetic relationships by constructing heavily engineered, domain-specific Knowledge Graphs \cite{song2026medspeak}. Despite yielding significant improvements in retrieval accuracy, these methods operate under a strict high-resource assumption. They require deep integration with the acoustic encoder's latent space, rely on massive, perfectly aligned speech-text training datasets, and demand immense computational overhead.

To address the heavy computational costs and limitations of classic retrieval, particularly for low-resource languages such as Persian, we propose a highly efficient and accurate Retrieval-Augmented Generation framework for ASR correction. Our approach introduces two vital innovations: an Error-Aware TF-IDF retrieval algorithm and a Symmetric Text Normalization module. Unlike methods that rely on N-best reranking, we use a single 1-best ASR hypothesis without requiring complex acoustic embeddings or knowledge graphs. By integrating our Error-Aware TF-IDF retrieval algorithm instead of standard TF-IDF, we reduce the word error rate from 21.95\% to 18.83\%, adding only a sparse matrix multiplication during inference. The main contributions of this letter are summarized as follows:

\begin{itemize}
    \item \textbf{Symmetric Text Normalization:} We introduce a robust preprocessing pipeline symmetrically applied to both the external knowledge base and online inference queries. This mitigates morphological mismatch and ASR loop hallucinations, preserving the mathematical integrity of the vector space.
    \item \textbf{Error-Aware TF-IDF Retrieval:} We modify the standard TF-IDF formula by calculating a weight for frequently hallucinated tokens, effectively forcing the system to retrieve targeted corrective context for the LLM \cite{salton1988term}.
    \item \textbf{Empirical Validation on Low-Resource Data:} Through extensive evaluation on the Persian Google FLEURS dataset using Whisper large-v3-turbo and Gemini 2.0 Flash, we introduce the Error-Aware Hit Rate (EA-HR) metric and demonstrate how each part of our approach affects the retrieval of related mistaken tokens \cite{conneau2023fleurs, radford2023robust}.
\end{itemize}

\section{Proposed Methodology}

First, as shown in Figure 1, our robust normalization module is applied symmetrically to the raw text during both KB construction and the online inference pipeline. This symmetric application ensures that loop hallucinations do not distort our probabilistic vector space. Furthermore, by standardizing multi-spelling words and formatting variations (for instance, converting numerical digits into consistent written words), we prevent token mismatches that could otherwise be misinterpreted by the system as ASR errors.

Second, we replace the standard TF-IDF retriever with our novel Error-Aware TF-IDF algorithm. Standard retrievers treat all tokens equally; however, our approach dynamically assigns distinct weights to correctly recognized words versus misunderstood tokens, forcing the system to retrieve documents that explicitly resolve the ASR model's phonetic failures. To achieve this, we calculate a sparse diagonal matrix of weights ($D_e$) based on the error propensity between our KB ground truth and the ASR predictions. During inference, this sparse matrix is simply multiplied by the standard TF-IDF matrix, ensuring there is no considerable computational latency compared to the baseline version. 

In the following subsections, we discuss the Symmetric Text Normalization and the new Error-Aware TF-IDF algorithm in detail.

\subsection{Symmetric Text Normalization}

The textual outputs generated by ASR systems often contain punctuation artifacts, zero-width non-joiner (ZWNJ) inconsistencies, and repetitive hallucinated tokens. To standardize the input space for the Information Retrieval (IR) module and make it clean for the LLM, a custom text processor is applied to both the ASR hypothesis and the RAG Knowledge Base. First, aligning with previous studies targeting Persian as a low-resource language, we normalize the ASR output by correcting spacing and ZWNJ characters, and converting all numerical digits into their standardized Persian formats. Eliminating these token mismatches has a vital impact on the error-aware retrieval process, which will be discussed in the following section.

Second, ASR models frequently exhibit loop hallucinations, endlessly repeating a previous token when they cannot recognize the subsequent acoustic signal. In some cases, the repeated word is entirely fabricated (e.g., continuously repeating a single word due to background noise). This severely degrades the RAG ASR correction pipeline in two ways. First, it adds unnecessary LLM token costs and context confusion. Second, it drastically skews term frequencies, rendering our TF-IDF vector space highly inaccurate. To address this, we truncate any token sequence repeating more than twice, discarding the excess tokens as hallucinations. This threshold serves as a tunable hyperparameter depending on the target language and domain.

\subsection{Error-Aware TF-IDF Retrieval}

Standard retrieval models, such as Term Frequency-Inverse Document Frequency (TF-IDF), have been utilized in previous studies to measure lexical similarity; however, these models inherently treat all tokens equally. In ASR post-processing with RAG, the goal of retrieval is to provide the Large Language Model (LLM) with explicit evidence to correct transcription errors. To illustrate this limitation, consider an ASR hypothesis comprising ten tokens, where two are phonetic misrecognitions. With standard TF-IDF, all tokens are treated equally. If the retrieval system finds a document containing only the eight correct tokens, it yields a high similarity score, despite offering zero corrective context to the language model for the two mistakes. Therefore, we must assign error-prone words significantly more weight than correctly recognized words. To address this, we propose an Error-Aware TF-IDF variant that dynamically recalibrates token weights based on their historical error probability.

We construct our Knowledge Base (KB) based on our ASR predictions and their respective actual targets after they have passed through our Symmetric Text Normalization. Let $V$ represent the vocabulary of the ASR predictions. For every term $t \in V$, we determine its ``error propensity'' by comparing the baseline ASR transcriptions against the ground-truth KB. Specifically, we tally the correct occurrences, $C(t)$, where $t$ is in the respective target of that row, and the error occurrences, $E(t)$, defined as instances where the ASR model hallucinated $t$ when it was not actually present in the ground truth.

To force the retrieval system to focus on these high-risk tokens, such as homophones and named entities, we introduce a dynamic weight multiplier, $W_{error}(t)$, calculated as:

\begin{equation}
W_{error}(t) = W_{c} + \left( \frac{E(t)}{E(t) + C(t)} \right) \times (W_{e} - W_{c})
\end{equation}

Here, $W_c$ and $W_e$ are empirically tuned hyperparameters. $W_c$ establishes the base weight for reliably recognized tokens, while $W_e$ dictates the maximum penalty weight applied to error-prone terms. In our experiments on the Google FLEURS dataset, we set $W_c = 0.1$ and $W_e = 2.0$. Consequently, tokens with a high historical hallucination rate significantly dominate the vector representation, ensuring they are not overlooked during document retrieval.

Using $W_{error}(t)$ for all $t \in V$, we construct a sparse diagonal penalty matrix $D_e$. During both KB vectorization and ASR query inference, this matrix is simply multiplied by the standard TF-IDF matrix:

\begin{equation}
\text{EA-TF-IDF} = \text{TF-IDF} \times D_e
\end{equation}

This architectural design guarantees two critical mathematical and computational advantages. First, it ensures that both the reference corpus and the incoming inference queries are symmetrically projected into a shared error-aware latent space, making downstream cosine similarity calculations mathematically sound. Second, because both TF-IDF and $D_e$ are highly sparse matrices, this multiplication adds near-zero computational latency during inference. This allows our framework to effectively bypass the massive processing overhead typically associated with other retrievers.

\section{Results and Experiments}

\subsection{Setup}

Considering Persian as a low-resource language, we utilize the Persian subset of the Google FLEURS dataset \cite{conneau2023fleurs} for evaluation. The training set, comprising 3,000 samples, is used to build the RAG knowledge base, and evaluation is run on the test set of 873 samples. For the ASR, we utilized Whisper large-v3-turbo on an RTX 3090. Furthermore, Google Gemini 2.0 Flash-Lite is used as the LLM to ensure cost-efficiency and low latency. 

During our observations of the Persian language, we identified errors related to optional connecting words and numerical representations. To prevent these inherent linguistic variations from skewing the final error rate calculations, we standardized the evaluation text.

To isolate the performance of the retrieval module, we designed the Error-Aware Hit Rate (EA-HR). Let the set of ASR hypothesis tokens for query $t$ in the knowledge base be $W(t)$ and its ground-truth token set be $G(t)$. The hallucinated error tokens for $t$ ($E(t)$) are defined by the relative complement:

\begin{equation}
E(t) = W(t) \setminus G(t)
\end{equation}

For an inference query represented by its ASR token set $Q_w$, we consider a retrieval successful if it retrieves at least one relevant hallucinated token:

\begin{equation}
E \cap Q_w \neq \emptyset
\end{equation}

\subsection{Results}

\begin{table}[t]
\caption{Offline Retrieval Evaluation (EA-HR @ Top-3)}
\label{tab:retrieval}
\centering
\begin{tabular}{lcc}
\toprule
\textbf{Retrieval Algorithm} & \textbf{Raw Text} & \textbf{Clean Text} \\
\midrule
Standard TF-IDF & 53.0\% & 53.7\% \\
\textbf{Error-Aware TF-IDF (Ours)} & \textbf{83.0\%} & \textbf{90.9\%} \\
\bottomrule
\end{tabular}
\end{table}

\subsubsection{Retrieval Evaluation}
First, we evaluate the performance of our proposed retrieval module by computing the EA-HR. This isolates its ability to identify the exact mistaken tokens within the query compared to standard TF-IDF. We tested both vectorizers on the identical Whisper ASR output from the Persian FLEURS test set. To measure the impact of Symmetric Text Normalization, we evaluated both retrievers with and without this preprocessing step. As shown in Table I, our Error-Aware TF-IDF achieves 83.0\% without normalization, compared to 53.0\% for standard TF-IDF. This demonstrates that integrating the error penalty significantly improves the retrieval of relevant mistaken tokens. Applying Symmetric Text Normalization further increased the EA-HR to 53.7\% for standard TF-IDF and 90.9\% for our method, highlighting the critical impact of morphological consistency.

\begin{table}[t]
\caption{End-to-End ASR Correction Performance}
\label{tab:wer}
\centering
\begin{tabular}{llc}
\toprule
\textbf{System Architecture} & \textbf{Retrieval Method} & \textbf{WER} \\
\midrule
Baseline ASR & None & 23.06\% \\
Standard RAG Correction & Standard TF-IDF & 21.95\% \\
\textbf{Error-Aware RAG (Ours)} & \textbf{Error-Aware TF-IDF} & \textbf{18.83\%} \\
\bottomrule
\end{tabular}
\end{table}

\subsubsection{End-to-End ASR Correction Evaluation}
The previous section demonstrated that Error-Aware TF-IDF successfully retrieves significantly more mistaken tokens than the standard method. This section evaluates its actual impact on the end-to-end RAG-ASR correction pipeline. To strictly isolate the performance of the retrieval modules, we test the full pipeline using both standard TF-IDF and Error-Aware TF-IDF, applying our Symmetric Text Normalization in both cases. As illustrated in Table II, the baseline ASR yields a 23.06\% Word Error Rate (WER), highlighting the inherent challenges of Persian ASR even after rigorous data cleaning. Implementing standard TF-IDF retrieval reduces the WER to 21.95\%. Ultimately, the proposed Error-Aware method achieves a superior WER of 18.83\%. These results confirm our hypothesis: the improved retrieval of mistaken tokens directly translates to enhanced LLM correction performance.

\section{Conclusion}

In this work, we explored methods to enhance ASR-RAG correction pipelines by providing more relevant augmented data, specifically targeting hallucinated tokens for LLM-based correction in low-resource languages. To achieve this, we introduced an Error-Aware TF-IDF retrieval algorithm and a Symmetric Text Normalization module. Experimental results demonstrate that these contributions significantly reduce the Word Error Rate (WER) without adding computational overhead, effectively addressing both data scarcity and inference latency. Future work will explore applying Error-Aware TF-IDF to domain-specific datasets and other RAG architectures. Additionally, we plan to evaluate the impact of expanding the knowledge base by running the baseline ASR model with diverse beam search parameters and temperatures to capture a broader range of potential hallucinations.

\section*{Acknowledgment}
The author wishes to thank Hamed Khayam Nekoei for his valuable discussions and support during the development of this work. Additionally, the author acknowledges that Gemini 3.1 Pro was utilized during the preparation of this manuscript strictly for language editing, grammatical correction, and LaTeX formatting to improve the clarity and readability of the human-authored text. The author assumes full responsibility for the final content and scientific integrity of this work.

\bibliographystyle{IEEEtran}
\bibliography{refs}

\end{document}